\documentclass[11pt]{article}
\usepackage{acl2014}
\usepackage{times}
\usepackage{url}
\usepackage{latexsym}

\usepackage[mathletters]{ucs}
\usepackage[utf8x]{inputenc}
\usepackage{amssymb}
\usepackage{amsmath}
\usepackage[usenames]{color}
\usepackage{hyperref}
\usepackage{wasysym}
\usepackage{graphicx}
\usepackage[normalem]{ulem}
\usepackage{caption}
\usepackage{subcaption}
\usepackage[english]{babel}
\usepackage[T1]{fontenc}

\usepackage{acronym}

\acrodef{SL}{Sign Language}
\acrodef{NLP}{Natural Language Processing}
\acrodef{FP}{False-Positive}
\acrodef{FN}{False-Negative}
\acrodef{MG}{Manual Gesture}
\acrodef{NMG}{Non-Manual Gesture}

\title{Sign Language Gibberish for syntactic parsing evaluation}

\author{R\'emi Dubot\\
       {IRIT}\\
       {Universit\'e de Toulouse}\\
       {\tt remi.dubot@irit.fr}\\\And
  Christophe Collet\\
       {IRIT}\\
       {Universit\'e de Toulouse}\\
       {\tt christophe.collet@irit.fr}\\
  }

\begin{document}

\maketitle
\begin{abstract}
\ac{SL} automatic processing slowly progresses bottom-up. 
The field has seen proposition to handle the video signal, to recognize and synthesize sub-lexical and lexical units. 
It starts to see the development of supra-lexical processing. 
But the recognition, at this level, lacks data. The syntax of \ac{SL} appears very specific as it uses massively the multiplicity of articulators and its access to the spatial dimensions. 
Therefore new parsing techniques are developed. However these need to be evaluated. 
The shortage on real data restrains the corpus-based models to small sizes. 
We propose here a solution to produce data-sets for the evaluation of parsers on the specific properties of \ac{SL}. 
The article first describes the general model used to generates dependency grammars and the phrase generation from these lasts. It then discusses the limits of approach. 
The solution shows to be of particular interest to evaluate the scalability of the techniques on big models. 
\end{abstract}
\section*{Introduction}
\begin{figure}
\centering
\includegraphics[width=\columnwidth]{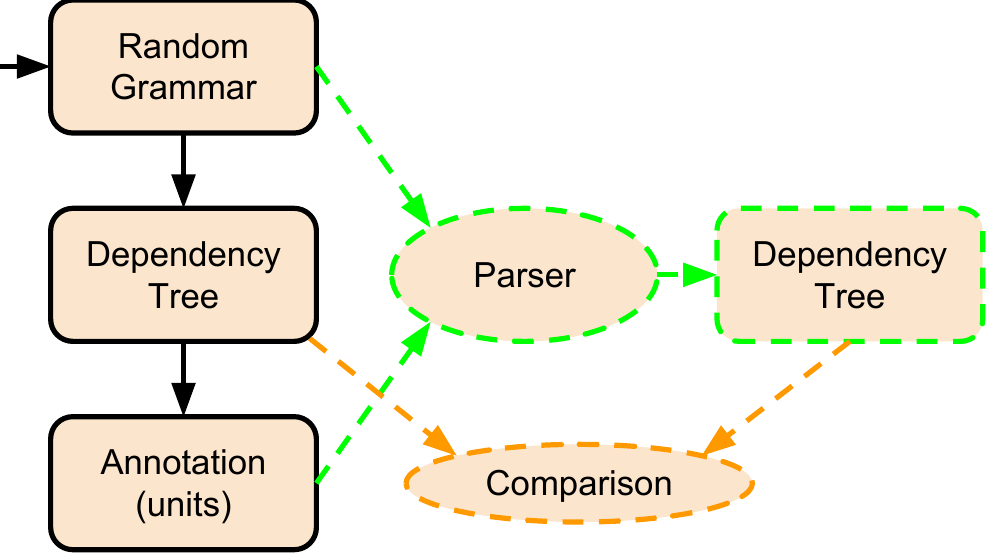}
\caption{General view of the targeted process}
\label{fig:gen-view}
\end{figure}

This article exposes a generative model reproducing syntactic properties of \acp{SL}. It addresses the generation of corpora of synthetic data for the evaluation of parsers against these properties. The focus is set on the generation of random grammars and corresponding syntactic trees (see Figure~\ref{fig:gen-view}). 

The syntactic analysis of \acp{SL} suffers from the absence of empirical data (corpus annotations) on the \acp{SL} syntax. This absence excludes the statistic approach to syntactic parsing. It should remain possible to use the knowledge extracted by \ac{SL} linguists. But, in practice, two problems arise. First, linguistic work generally lacks formally annotated examples. This makes the informal descriptions of phenomena difficult to interpret and formalize. Second, the evaluation of syntactic analysis still requires annotations to use as a ground truth.
Only a joint work between NLP researchers and linguists can lead to the required formal descriptions and annotation of syntactic phenomena.
The proposed corpus generation produces ground truth for parser evaluations.

This document is structured as follows. It begins with a presentation of related work on synthesis of \ac{SL} and vocal languages. The description of our contribution starts with the specification of the considered \ac{SL} properties. Then we introduce syntactic tree and grammar generation. Finally we discuss the use of the produced corpora in evaluations. 

\subsection*{Related work}
\ac{SL} synthesis is not a new field of study but its syntactic level has been ignored. Most works on synthesis are centered on avatar animation~\cite{kipp_sign_2011}. They use as input either motion capture data or morphological level descriptions (e.g. Hamnosys~\cite{hanke_hamnosys-representing_2004}). 
As an exception, the formalism created by Filhol has been demonstrated on structures from sub-lexical to supra-syntactic~\cite{filhol_combining_2012}. It describes the minimal constraints in structures that make them recognizable. However, the structures are described independently and no relation can be expressed between them. Consequently, it only addressed synthesis on the surface ; the composition of the structures to build a locution has to be done by hand.
None of these solutions allow to automatically produce a synthetic corpus annotated in syntax.

For vocal languages, the corpus-based analysis started in early 90’s, at this time the field already had mature theories and a large community. In consequence, corpora were already present or ready to constitute. Corpus synthesis has been used to generate precisely annotated corpora at signal level for the voice (as well as for music). We cannot find any trace of corpus synthesis for the syntactic level.  

\section{Specification}

The purpose of this work is to produce data syntactically similar to \ac{SL}. It mimics the synchronized parallel unit production and the possibly weak word order. We consider these properties as the most difficult to deal with for syntactic parsers. 

%\begin{figure}
%\centering
%\includegraphics[scale=.8]{"Gen Org View"}
%\caption{General view of the targeted process}
%\label{fig:gen-view}
%\end{figure}

\subsection{Objectives}

First point, \acp{SL} use multiple articulators to produce a high dimensional signal. \ac{SL} linguists concur to say that one can decompose this signal into several units produced in parallel. The relations between these units show complex synchronization mechanisms~\cite{chetelat_les_2010}. These sort of relations are described,  for example, in modal markers, syntactic markers, references, qualifiers, quantifiers, etc. This is a real difficulty for the syntactic parsers developed for vocal languages as they are based on sequences\footnote{The reader should notice this problem is independent of the word order strictness. }. It is the reason why, the generated data should represent these sort of structures. 
Second point, the strictness of the word order is still controversial. Neidle \textit{et al}.~\cite{neidle_syntax_2000} support the strict word order when Cuxac \textit{et al}.~\cite{cuxac_langue_2000} and Dubuisson \textit{et al}.~\cite{dubuisson_grammaire_1999} support weak word order. We want a model able to generate data for these two hypotheses. 

\subsection{Simplifications}
The model makes several simplistic hypotheses. 
It divides the units in two types: \acp{MG} and \acp{NMG}. Each has its proper behavior. The units can represent as well standard signs, other \acp{MG} (e.g. pointing \acp{MG}), facial gestures (e.g. qualifiers, quantifiers, modality markers), gaze gestures (e.g. references), etc. 

In \acp{SL}, articulatory constraints impact the syntactic level. 
%This is a generalization of the co-articulation:  
Some units interact and some others are incompatible. 
The model emulates simplified articulatory interactions between its units:  
\begin{itemize}
\item \acp{MG} cannot overlap. This excludes the representation of described phenomena (e.g. buoy structures, Cuxac's situational-transfers). 
\item \acp{NMG} always can overlap. This is a simplification as some \acp{NMG} are articulatorily impossible to produce simultaneously. 
\end{itemize} 
These simplifications appear to be compatible with the presented objectives. 

\section{Annotation generation}
\begin{figure}
\centering
\includegraphics[width=\columnwidth]{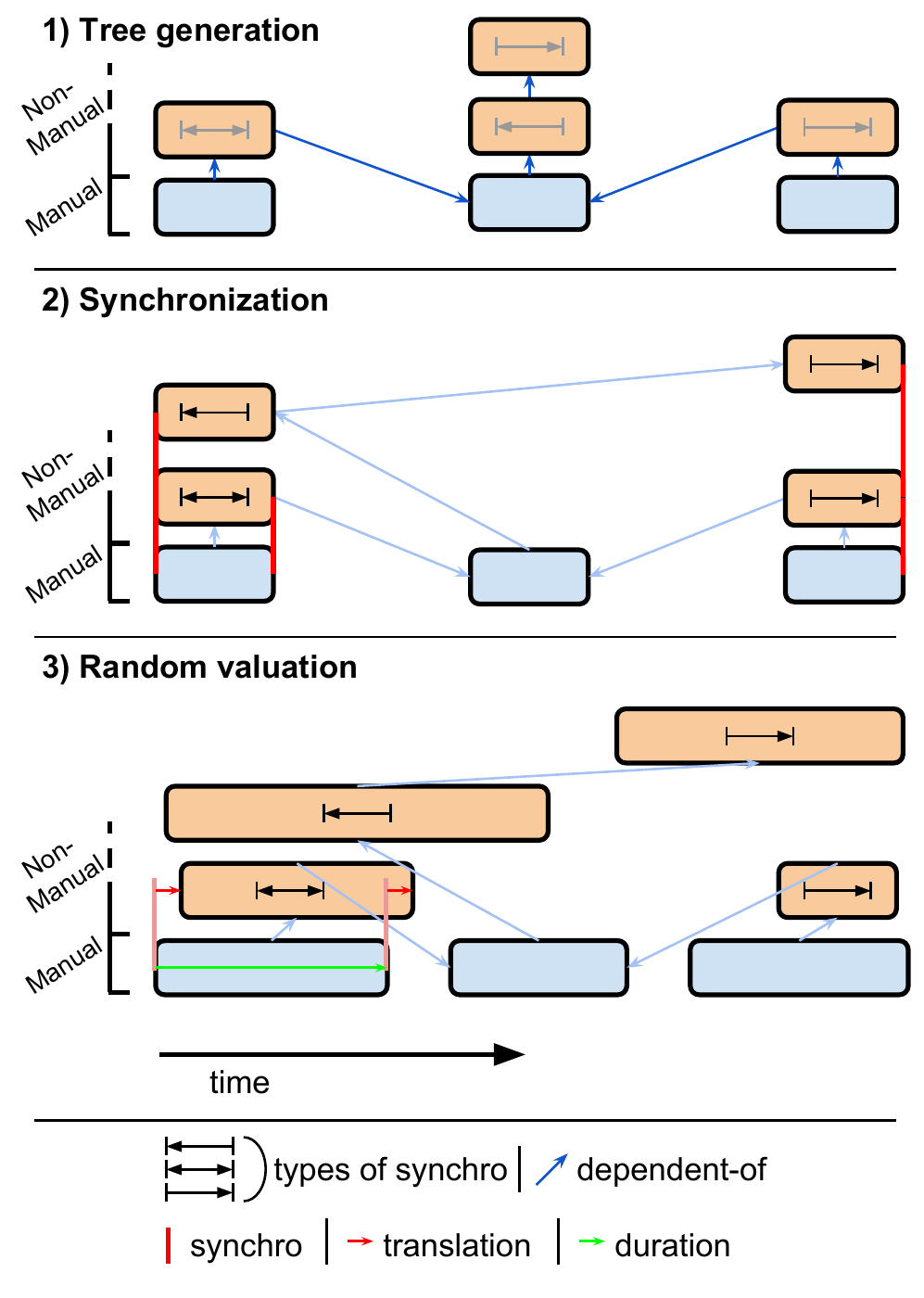}
\caption{Dependency tree production step-by-step}
\label{fig:ex-prod}
\end{figure}

The model uses dependency grammars to produce dependency trees. The targeted result is shown in figure~\ref{fig:ex-prod}. 
The synthetic annotations are made in two steps: the generation of the dependency tree and the valuation of the temporal attributes. 

%\begin{figure}
%\centering
%\includegraphics[width=\columnwidth]{"DepTree Ex steps"}
%\caption{Dependency tree production step-by-step}
%\label{fig:ex-prod}
%\end{figure}

%\begin{figure}
%	\centering
%    \includegraphics[scale=.8]{"DepTree Ex var"}
%    \caption{Example of result}
%    \label{fig:ex-prod}
%\end{figure}
%
%\begin{figure}
%	\centering
%    \begin{subfigure}[b]{\columnwidth}
%	    \caption{Synchronization constraints}
%	    \label{fig:sync}
%	    \includegraphics[scale=.8]{"DepTree Ex sync var"}
%    \end{subfigure}\\
%    
%    \begin{subfigure}[b]{\columnwidth}
%        \caption{Valuation}
%        \label{fig:trans}
%        \includegraphics[scale=.8]{"DepTree Ex trans var"}
%    \end{subfigure}\\
%    
%
%\caption{Dependency tree production step-by-step}
%\label{fig:steps}
%\end{figure}

\subsection{Tree generation}
We base our work on Hays' dependency grammars. It defines rules of the form $X(Y_{-n},...,Y_{-1},*,Y_1,...,Y_m)$ where $X$ and $Y_k$ are categories of units. Such a rule expresses that a unit of category $X$ takes the place of the star in a sequence of dependents of categories $Y_{-n}$ to $Y_m$. 
This formalism is enough for \acp{MG} (assuming the sequence simplification). 
The \acp{NMG} requires to extend it with rules of the form $X(Y)$. 

The generation of a random tree works like usual derivational grammars. As a consequence, it needs a root category. The rules are chosen following an heuristic to ensure convergence. In a final step, the categories are replaced by units. This process leads to a tree similar to the one in the first part of Figure~\ref{fig:ex-prod}.

The \acp{NMG} are all considered as markers. This is again a simplification. As shown on figure~\ref{fig:ex-prod}, the \acp{NMG} affect their unique dependent. 
The literature of \ac{SL} linguistics gives examples of markers maintained all along the projection of their dependent or emitted at the beginning or at the end of the projection. 
The figure~\ref{fig:ex-prod} shows these three types of \acp{NMG}. 

\subsection{Temporal valuation}

Temporal attribute values are assigned in two steps. First, the relative positions (translation) and durations are picked at random. Second, the absolute positions are computed by solving a constraint network. 

\begin{figure}
\centering
\includegraphics[width=\columnwidth]{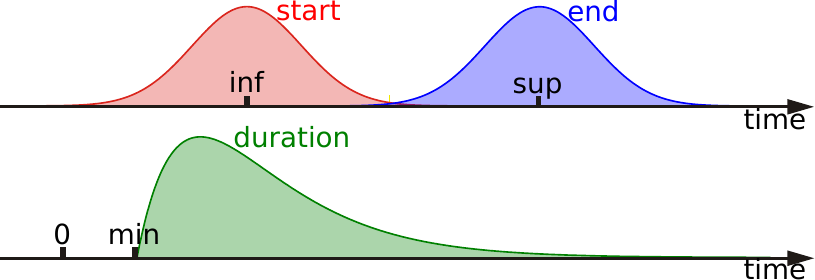}
\caption{Start/End distributions}
\label{fig:start-end}
\end{figure}

In absence of measures on annotated corpora, the distribution of the values (\ac{MG} duration and \ac{NMG} translations) are chosen arbitrarily. As shown in figure~\ref{fig:start-end}, the start and the end are translated following Gaussian distributions. When only one end is synchronized, the duration is randomly chosen from a gamma distribution.

The absolute positions are computed using a CSP where nodes are units with their two temporal attributes and the constraints are the durations (of \acp{MG}), the sequence (between \acp{MG}) and the synchronizations (of \acp{NMG}, with translations, see Figure~\ref{fig:ex-prod}).  

\section{Generation of grammars}
The previous part showed how to produce random data from grammars. This part addresses the generation of random grammars. The grammars include the following elements: categories, rules and units. The categories are containers for rules and units. Their production is straightforward. 
%However, the next part discusses the impact of the number of units and rules per category. 
This part focuses on the rules and units production. 

\subsection{Rule generation}
The rule generation is parametrized with several probabilistic distributions. There are distributions for the number of rules per category and the number and positions of the dependents in a rule. 

A weak word order is emulated by introducing rules which are permutations of the elements of existing rules. The strictness is adjusted with the probability of introduction of permutations. 

The generated grammars must be finite. It means that starting from any category, the grammar must produce at least one finite dependency tree. 
Prolog gives one solution to verify this property. Indeed, the Hays' rules can be translated to Horn clauses. In this case, a category is finite if, and only if, the corresponding proposition is a theorem. The finiteness of the grammar is then achieved by either removing non-finite categories or introducing new rules. 
However, we want to further the finiteness. We introduce a metric: the height of the grammar. 
With the proper definitions, the finiteness of a grammar or a category is equivalent to the finiteness of its height. 
The strategy used to limit the height is iterative: find the categories which have a height exactly equal to the limit ; add a leaf (empty) rule to them.  
 
\subsection{Unit generation}
The unit production requires several parameters but does not pose particular problems. 
The choice between a \ac{MG} unit and one of the three types of \ac{NMG} units is random with a fixed ratio. 
As stated before, the units use parametrized distribution for the data generation. In our implementation: all the gamma distributions have a shape parameter of 2, the durations take a scale parameter in a normal distribution. 

\section{Discussion}
The model description has shown many parameters. 
They are generally described and implemented as independent probability distributions. 
Current statistic knowledge of \acs{SL} does not help to tweak the model's distributions.
A (even small) dependency annotation on a corpus of real \ac{SL} would allow to refine the approximations of some parameters (e.g. durations, position of the head). 

However, some parameters are strongly dependent of the theoretical framework. 
It the case of the choice of the number of rules per category $R_C$ and the number of units per category $U_C$. This impacts the type of grammar produced. Favoring $R_C$ over $U_C$ produces grammars closer to what would produce an automatically learned grammar. On the contrary, favoring $U_C$ over $R_C$ produces highly regular grammars closer to what a linguist would produce. 
Generally, solving these questions implies a level annotations and analysis not reachable in a near future. 

The model generates ambiguous grammars. For example, two rules $A(*,B)$ and $B(A,*)$ in a same grammar introduce ambiguity. Half-synchronized markers do the same. We consider ambiguity as part of natural languages. Consequently, the parsers should be robust to ambiguity and this ability must be evaluated. %on the presence of the solution in the output set (with the usual set of measures: FP, FN, precision, recall, etc.). 

\section*{Conclusion}
This paper propose a corpus synthesis which mimics syntactic features of \acp{SL}. %It focus on features which are problematic for parsers. 
This synthetic data is necessary to enable the work on parsers. Indeed, we know that --by design-- the current parsing solutions do not suit \acp{SL} but the lack of empirical data restrain their adaptation or the development of new solutions. 
The reproduced features are the ones which are problematic for parsers, namely: the production of units with synchronization mechanism, and the discussed strictness of the word order. 
We presented the process from the generation of random dependency grammars to the generation of random dependency trees.  
The model's parameters enable the characterization of the parsers by running them on multiple parameter sets. %The evaluation on ambiguous grammars requires the parsers to produce a set of output rather than a single output. 

The solution can be improved in several directions. The approximation of the model's parameters would benefit from a dependency annotation. As presented, the solution simulates a perfect unit recognition layer. Adding noise enables the robustness evaluation. Noise can affect both annotations and grammars. The model itself can also be refined. It can take into account the finite number of articulators and uncouple the two hands. In this case, it would need other constraints (articulatory and grammatical). The model can also integrate the spatial dimension of \acp{SL}, allowing to evaluate parsers on the spatial reference resolution. The development is guided by the evolution on the parsers. 
To conclude, we believe that this work does not reduce the urgent need for massive syntactic annotation of \acp{SL} but still unblock the situation.

\bibliographystyle{acl}
\bibliography{MyLibrary}

\end{document}